# A Community Based Algorithm for Large Scale Web Service Composition

[1]Chantal Cherifi, [2]Yvan Rivierre, [3]Jean-François Santucci

[*1]SPE Laboratory, *UMR 6134, University of Corsica, France, chantalbonner@gmail.com*
[2]*VERIMAG UMR 5104, J. Fourier University, Grenoble, France, Yvan.Rivierre@imag.fr*
[3]SPE Laboratory, *UMR, 6134 University of Corsica, France, santucci@univ-corse.fr*

## Abstract

*Web service composition is the process of synthesizing a new composite service using a set of available Web services in order to satisfy a client request that cannot be treated by any available Web services. The Web services space is a dynamic environment characterized by a huge number of elements. Furthermore, many Web services are offering similar functionalities. In this paper we propose a model for Web service composition designed to address the scale effect and the redundancy issue. The Web services space is represented by a two-layered network architecture. A concrete similarity network layer organizes the Web services operations into communities of functionally similar operations. An abstract interaction network layer represents the composition relationships between the sets of communities. Composition synthesis is performed by a two-phased graph search algorithm. First, the interaction network is mined in order to discover abstract solutions to the request goal. Then, the abstract compositions are instantiated with concrete operations selected from the similarity network. This strategy allows an efficient exploration of the Web services space. Furthermore, operations grouped in a community can be easily substituted if necessary during the composition's synthesis's process.*

**Keywords**: *Web services, composition, substitution, interaction, similarity, community, network, graph search algorithm*

## 1. Introduction

The paradigm of Web service is gaining more and more popularity with companies and organizations that are interested in lowering the cost of the development and maintaining of their applications. Indeed, Web services are software systems designed to support interoperable machine-to-machine interaction over a network. They allow enterprises to implement their core business as services over the Internet. Once published in a registry by providers, they can be discovered and invoked by business partners or clients. In order to achieve new and more useful functionalities, they can be programmatically loosely coupled through the Web. Resulting value-added composite Web services can satisfy a user request when no atomic Web service is able to do it. Automatic and dynamic composition process nevertheless raises interesting challenges. Among them is the scale effect; Web services are numerous on the Web and their number is increasing with time. Besides, they are created, changed, relocated, or even removed on the fly; this volatile aspect is another source of complexity. Another particularity is that lot of them provide overlapping or identical functionalities with eventually different quality of service. This results in a huge, intricate and dynamic space to be explored. The topic always stirs researcher interest and various propositions have been made to meet the challenges.

Many proposals address Web service composition as a planning problem [1], [2]. With the increasing number of available Web services, such solutions suffer from their high complexity and a prohibitive computational cost. Other approaches treat the composition as a graph search problem. Indeed, in the composition context, the Web services space can naturally be represented by a network of interacting atomic Web services. In [3], compositions are discovered within a semantic Web service network by a forward chaining algorithm. In [4], search of compositions in a syntactic Web service network is performed using graph matching techniques. In [5], a semantic Web service network is stored in a relational database. Composition search is done by SQL statements. In [6], a breadth first search algorithm is used to search for compositions in a semantic network of parameters. In [7], the information on link analysis of a semantic Web service interaction network is used to guide the A* shortest path search algorithm that probe the Web service space. The authors in [8], use the A* search algorithm to find a minimal composition within a subset of semantic Web services represented as a graph. The approach in [9] proposed the use of a semantic interaction network enriched with an organization of the Web services in communities. This is an ontological organization where communities are sets of Web services providing services in the same domain. The network is used to search for compositions through a forward chaining algorithm. In [10], the authors propose a dynamic Web service composition algorithm based on the combination of ant colony algorithm and genetic algorithm, to address the efficiency issue in a large solution space. In [11], genetic algorithms allow quality of service-aware Web service composition. The authors in [12] propose a framework to deal with data distribution and quality of service issues by solving problems of unavailability of updated information and inaccessibility of Web services. Note that there is a great deal of work addressing the composition issue not only according to the Web services functional requirements, but also to their transactional properties their QoS characteristics or the security problems [13], [14].

In current solutions, discovery and composition processes take place within a predefined space which is a repository of individual and atomic Web services. Even, if they can solve some key issues in Web service composition, none of them gives an effective solution to address the scaling and redundancy effect. In this paper, we present a graph search based approach to overcome these shortcomings. We propose to use a two-layered network architecture to store interaction and similarity functional relationships that occur between the operations of Web services. Note that we consider operations rather than Web services as atomic element because it is at this level that Web services are used. In order to reach a request goal, a composition search algorithm explores the two network layers in two pass. First, an "abstract interaction network" is mined in order to retrieve a set of meta-operations that satisfies a given goal. Then, the meta-operations are instantiated with real operations extracted from a similarity network. The similarity network gives opportunities to substitute operations that offer similar functionalities. This need is susceptible to happen either when a user is not satisfied with non functional aspects of a Web service, i.e. quality of service, or when a Web service is out of service for different reasons. The interaction layer is built above the similarity layer. Similar operations are grouped into a single meta-operation in order to reduce the search space during the first phase of the algorithm. The main features of our proposal are that it allows:
1) Reducing the search space by using similarity between Web services functionalities.
2) Taking advantage of this similarity to give opportunities to substitute Web services and to efficiently deal with redundancy.

The rest of this paper is organized as follows. In section 2, we present the two-layered network architecture and we describe the abstract layer and the instance layer. Section 3 is devoted to our composition algorithm with details on the two phases. Finally, we end the article in section 4 by discussing some conclusions and directions for future work.

## 2. The two-layered architecture

Our approach for Web service composition is based on a structure of two networks. An abstract interaction network is used to discover a set of meta-operations that can potentially satisfy a goal. A concrete similarity network allows to instantiate them with actual and available Web service operations. Groups of similar operations are pre-existing structures stored in a network of similar concrete operations. Composition of meta-operations is a pre-existing structure stored in a network of interacting meta-operations. They can be upgraded easily when new Web services appear.

### 2.1. Instance layer

The instance layer represents all concrete operations of published Web services. They can be potentially invoked to fulfill a request. They are organized in a similarity network.

A similarity network of operations is a graph whose nodes correspond to Web service operations and links indicate that two operations offer similar functionalities. As our main concern is to deal with Web services substitution, we consider that two operations are similar if they allow reaching more or less the same goal. Hence, to determine the similarity, we consider input and output parameters. We consider four similarity levels called *Full Similarity*, *Partial Similarity*, *Excess Similarity* and *Relation Similarity* [15]. They are defined in terms of set relations between the input and the output parameter sets of the compared operations. Suppose we want to compare two operations $o_1$ and $o_2$. Let $I_i$ be the set of input parameters, and $O_i$ the set of output parameters for operation $o_i$. A *FullSim* network is obtained using a symmetrical function such that two operations are fully similar if and only if 1) they provide exactly the same outputs ($O_1 = O_2$) and 2) they need overlapping inputs ($I_1 \cap I_2 \neq \emptyset$). *PartialSim* and *ExcessSim* networks are associated to asymmetrical functions. In the former, $o_2$ is partially similar to $o_1$ if and only if 1) some $o_1$ outputs are missing in $o_2$ ($O_1 \supset O_2$) and 2) they need overlapping inputs ($I_1 \cap I_2 \neq \emptyset$). In the latter, $o_2$ is similar to $o_1$ with excess if and only if 1) $o_2$ provides all $o_1$ outputs plus additional ones ($O_1 \subset O_2$) and 2) $o_2$ needs only some of $o_1$ inputs ($I_1 \supseteq I_2$). A *RelationSim* network uses a symmetrical function. Two operations have a relational similarity if and only if 1) they have exactly the same outputs ($O_1 = O_2$) and 2) they do not share common input ($I_1 \cap I_2 = \emptyset$).

In the following, the instance layer is realized by a FullSim network of operations. For short, we will refer to it as "similarity network of operations". It is the most satisfying level of similarity from a substitution point of view. Nevertheless, even if they offer a less effective solution, the other networks can be also considered. All the similarity networks exhibit a component structure [15]. A component is a maximal connected sub-graph i.e. a set of interconnected nodes, all disconnected from the rest of the network. Each component materializes a community, i.e. a group of similar operations. The lower part of Figure 1 represents the communities extracted from a set of 8 operations. There are four communities represented by different colors. The operations with the same color belong to the same community. Note that communities are non-overlapping and include all the network nodes.

A remarkable structure within the components is the clique. A clique is a fully connected sub-network. In a similarity network of operations, a clique contains operations that share at least a common input parameter. In the lower part of Figure 1, operations ($o_2$, $o_3$, $o_4$) form the largest clique of the network, with b as common parameter. Operations in a same clique are the most likely to be substituted.

### 2.2. Meta-layer

The meta-layer enables to search for compositions in a reduced space of an interaction network of meta-operations. A meta-operation is the representative of a community in the instance layer. We define a set of input parameters and a set of output parameters for each meta-operation. The set of input parameters of a meta-operation is defined as the union of the inputs of all the operations of the corresponding community. Similarly, the output parameter set of a meta-operation is the union of the output parameter set of the operations of the underlying community. Meta-operations are linked together to form an interaction network of meta-operations.

An interaction network of meta-operations is a directed graph *N (V, E)*, where *V* is the set of nodes representing the meta-operations and *E* is the set of links representing their interactions. Let two meta-operations $m_i$ and $m_j \in V$, there is a directed link $(m_i, m_j) \in E$, if and only if $m_i$ can interact with $m_j$. Meta-operations interact according to the partial invocation mode. In other words, a meta-operation $m_i$ can interact with a meta-operation $m_j$, if and only if $m_i$ has at least one output parameter which is similar to one of the input parameters of $m_j$.

Figure 1, illustrates this two-layered architecture. Meta-operations are represented with rounded-corner boxes, above the underlying connected components (communities) of the similarity network. By convention, all notations related to the meta-layer are written in *cursive* script, in order to distinguish it from the rest of the model.

The upper part of Figure 1 represents an interaction network with four meta-operations, $m_1$, $m_2$, $m_3$ and $m_4$, extracted from the four components of the similarity network illustrated in the lower part of the figure. Operations $o_1$, $o_2$, $o_3$ and $o_4$ are represented by the meta-operation $m_1$, operations $o_5$ and $o_6$ are represented by the meta-operation $m_2$, and operations $o_7$ and $o_8$ are represented by meta-operations $m_3$ and $m_4$ respectively. Links are labeled with the set of parameters that enable the invoking meta-operation to interact with the invoked meta-operation. $m_1$ interacts with $m_2$ trough parameter e, $m_3$ interacts with $m_2$ trough parameter e and $m_4$ interacts with $m_3$ trough parameter f.

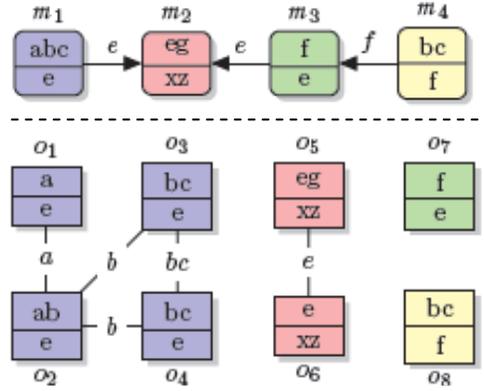

**Figure 1.** Architecture of the two-layered network model for Web services composition. Down: Full similarity network representing the instance layer; similar operations are grouped into communities (same color). Up: Partial interaction network representing the meta-layer; a meta-operation represents a community of the instance layer.

## 3. Composition algorithm

The composition process consists of two steps. In the first step, the meta-operation network is explored to search for a meta-composition that fits a user request. It is performed using a graph-based approach, starting from the goal of a service's request and composing backwards in the direction of the input of the service request. This meta-composition wraps all possible compositions. If no meta-composition is found, it implies that no composition could have been found anyway. In the second step, the meta-composition is instantiated into compositions of operations. By instantiating we mean replacing the abstract meta-operations by concrete operations of the similarity network of the instance layer. A composition of operations is a sequence of operations sorted by invocation order. It does not contain the functional relations between these operations, but they can be easily inferred. Note that the sequence is only used for simplifying the formalism. Even if formally expressed as a sequence of operations, composition can be either sequential, parallel dependent or parallel independent.

The algorithm can find all the compositions that fit a user request. Depending on its needs and as long as he is not satisfied by the returned composition, a user can call the second step several times, in order to obtain other possible compositions. It is a valuable behavior, because retrieving all possible compositions can be very expensive. We assume that the user is interested in getting a suitable composition without missing an opportunity of composition, rather than getting all possible ones.

### 3.1. Environment

This environment is defined for formalizing both steps of the general composition algorithm. K is the set of parameters known by the user, i.e. the input part of the request. G is the set of desired parameters, i.e. the goal part of the request. N = (V, E) is the operation similarity network. $\mathcal{N}$ = (V, E) is the meta-operation interaction network built on top of N. The following primitives are used to explore a network. `Label`(m, m') returns the label of edge (m, m') in N. `Input`(m) returns the input parameters of m. `Output`(m) returns the output parameters of m. `Pred`(m, *network*) returns the predecessors of m in *network*. Since a meta-composition is a sub-network of N, it can be considered as a network as well. Finally, the two kinds of the results provided by the general composition algorithm are stored into variables declared as follows. *C* is the meta-composition to be built at first step, and instantiated at second step. This is a network of meta-operations. C is the composition to be built at second step, by instantiating the meta-composition *C*. This is a sequence of operations. The two steps of the algorithm are detailed in the following sub-sections.

### 3.2. Find Meta-composition

Find meta-composition is the first step of the general composition algorithm. It consists of finding a meta-composition *C* which fits the request. The goal G must be supplied by meta-operations of *C*. Only K, that contains the parameters known by the user, can be used. This step is called exactly one time per user request. If there is no meta-composition which can fulfill the whole goal, false is returned.

The first step of the general composition algorithm is formalized in Procedure 1. First, meta-operations which supply at least a part of the goal G are listed as candidate. They serve as entry points to begin the *N* network exploration (line 4: build initial candidate paths). Then, a backward search is performed from each of these starting points as follows. When visiting a meta-operation *m* from a specific path, there are two possibilities for adding path meta-operations to *C*. Either at least one input parameter of *m* is known by the user (line 9) or *m* has at least a predecessor of *m* that is already in *C* (line 12: path head is already in meta-composition). To pursue the exploration from *m*, incident links are followed if and only if they provide at least an unknown parameter (line 15) and they don't lead to an already visited meta-operation (line 1: at least a parameter can be obtained from *m* and no cycle is created). The latest condition ensures that the exploration does not follow cyclic paths and eventually terminates. Finally it verifies (line 21: effectively supplied parts of the goal) that each part of the goal G can be provided by at least a meta-operation of the resulting meta-composition *C*. If it is effectively the case, it successfully returns *C*. Otherwise it returns false as an indication that no composition can fulfill the request.

The first step of the general composition algorithm depends on the declarations below. The following variables are locally declared for this step. *Path* is the currently visited path of meta-operations in *N*. *Next* is the set of candidate paths for further iterations. The primitives hereafter are used in this step only. Add(*path, meta-composition*) adds meta-operations of *path* to *meta-composition*. Head(*path*) returns the head element of *path*. Pop(*set of paths*) returns and removes a candidate *path* from *set of paths*. Suppliers(*parameters, network*) returns meta-operations in *network* which supply at least one of *parameters*. This primitive is used to search for the goal.

After this step, the meta-operations interaction network *N* is no more considered in its integrity. Only its sub-network defined by the meta-composition *C* stands in scope of the second step of the general composition algorithm.

### 3.3. Instantiate Meta-composition

The second step of the general composition algorithm consists in instantiating the meta-composition *C* found at the first step. This step can be called several times. At each call of the instantiation step, another composition is built from the same meta-composition until no more composition can be found.

The second step of the general composition algorithm is formalized in Procedure 2. The sets of meta-operations that cover the goal are processed one after the other. Those meta-operations are the starting points for the similarity network exploration. After this initialization (line 1 and 2: at first use, (re)initialize the set of meta-operations used to reach the goal) the network is explored for a next instantiation setup (line 4), either by instantiating a meta-operation *m* to a different operation, or by using another set of input-covering meta-operations for *m*. If there are no more possible instantiation for this set of goal-covering meta-operations, then (line 11) the next set of goal-covering meta-operations is selected. If there are no goal-covering sets left, it means that all instances have been already returned by previous call to this instantiation procedure. Finally, the network is explored (line 19), without modification, for extracting and returning the selected instance of the meta-composition.

The last step of the general composition algorithm depends on the declarations below. At each call of this step, the following local variables are declared. *HasNext* returns true if a meta-operation has another instance or covering set to be used next, false else. *Visited* is the set of meta-operations that have been visited before. It is used in order to avoid infinite recursion.

The meta-composition $C$ is persistently backed with additional variables, in order to select the appropriate nodes and links of the meta-composition. One of these variables is used to iterate over meta-operations sets that cover the goal. There are two additional variables per meta-operation. The first one is iterating over invocable instances of a meta-operation. The second variable iterates over meta-operations sets that cover input of its instance. All of these variables can exclusively be used through the following primitives. GetGoalCover() returns the selected set of meta-operations to cover the goal. If none is selected, false is returned. Initially, no set is selected. NextGoalCover() selects and returns the next set of meta-operations to cover the goal. If there is none, it deselects the currently selected set and returns false. GetInstance($m$) returns the selected instance for meta-operation $m$. If none is selected, false is returned. Initially no instance is selected. NextInstance($m$) selects and returns the next invocable instance for meta-operation $m$. If there is none, it deselects the currently selected instance and returns false. It also deselects the currently selected cover. GetInCover($m$) returns the selected set of meta-operations that covers input of the selected instance for meta-operation $m$. If there is none, false is returned. Initially no set is selected. NextInCover($m$) selects and returns the next set of meta-operations that covers input of the selected instance for meta-operation $m$. If there is none, it deselects the currently selected set and returns false.

At the end, two subroutines are defined in Procedure 3 and Procedure 4, in order to handle recursive search through the meta-composition network $C$. GetOpSeq($m$) returns a sequence of operations which instantiates the sub-network of $C$ which ends to $m$. It is recursively defined and uses the *Visited* set of meta-operations in order to avoid infinite recursion. NextOpSeq($m$) returns true if there is a next possible instantiation of the sub-network of $C$ which ends to $m$. It is recursively defined and uses the *Visited* set of meta-operations in order to avoid infinite recursion.

| **Procedure** 1. Find meta-composition | **Procedure** 2. Instantiate meta-composition |
|---|---|
| 1: $C \leftarrow \emptyset$ | 1: **if** GetCoverGoal( ) = $\bot$ **then** |
| 2: $Next \leftarrow \emptyset$ | 2:   NextCoverGoal( ) |
| 3: **for all** $m \in$ Suppliers(G, N) **do** | 3: **end if** |
| 4:   $Next \leftarrow Next \cup \{\{m\}\}$ | 4: $HasNext \leftarrow Next \bot$ |
| 5: **end for** | 5: $Visited \leftarrow Next \emptyset$ |
| 6: **while** $Next \neq \emptyset$ **do** | 6: **for all** $m \in$ GetCoverGoal( ) **do** |
| 7:   $Path \leftarrow$ Pop($Next$) | 7:   **if** $HasNext = \bot$ **then** |
| 8:   **if** Input(Head($Path$)) $\cap K \neq \emptyset$ **then** | 8:     $NasNext \leftarrow$ NextOpSeq($m$) |
| 9:     Add($Path, C$) | 9:   **end if** |
| 10:  **end if** | 10: **end for** |
| 11:  **if** Head($Path$) $\in C$ **then** | 11: **if** $HasNext = \bot$ **then** |
| 12:    Add($Path, C$) | 12:   NextCoverGoal( ) |
| 13: **else** | 13:   **if** GetCoverGoal( ) = $\bot$ **then** |
| 14: **for all** $m \in$ Pred(Head($Path$),$N$) **do** | 14:     **return** $\bot$ |
| 15: **if**(Label(m,Head($Path$))\\$K \neq \emptyset$) $\wedge$ (m$\notin Path$) **then** | 15:   **else** |
| | 16:     $HasNext \leftarrow \top$ |
| 16: $Next \leftarrow Next \cup \{\{m\}.Path\}$ | 17:   **end if** |
| 17:     **end if** | 18: **end if** |
| 18:   **end for** | 19: $C \leftarrow \emptyset$ |
| 19:  **end if** | 20: $Visited \leftarrow \emptyset$ |
| 20: **end while** | 21: **for all** m $\in$ GetCoverGoal( ) **do** |
| 21: $Supplied \leftarrow \emptyset$ | 22:   **if** GetInstance($m$) $\neq C$ **then** |
| 22: **for all** $m \in Suppliers$($G, N$) **do** | 23:     $C \leftarrow$ NextOpSeq($m$).$C$ |
| 23:  $Supplied \leftarrow Supplied \cup$ Output($m$) | 24:   **end if** |
| 24: **end for** | 25: **end for** |
| 25: **if** $G \subseteq Supplied$ **then** | 26: **return** $C$ |
| 26:   **return** $C$ | |
| 27: **else** | |
| 28:   **return** $\bot$ | |
| 29: **end if** | |

| **Procedure 3.** Instanciation Subroutine | **Procedure 4.** Iteration Over Instances |
|---|---|
| 1: **procedure** GetOpSeq(*m*)<br>/returns operations sequence for subnetwork of *C* ending with *m*/<br>2:    Visited ← Visited ∪ {*m*}<br>3:    **if** GetInstance(*m*) = ⊥ **then**<br>      /switch to first instanciation setup of *m*/<br>4:        NextInstance(*m*)<br>5:        NextInCover(*m*)<br>6:    **end if**<br>7:    OpSeq ← {GetInstance(*m*)}<br>/initialize operations sequence with *m*/<br>8:    **for all** *m'* ∉ Visited ∧ *m'* ∈ GetInCover(*m*) **do**<br>9:        OpSeq ← GetOpSeq(*m'*).OpSeq /prefix operations sequence with the one of *m'*/<br>10:   **end for**<br>11:   **return** *OpSeq*<br>12: **end procedure** | 1: **procedure** NextOpSeq(*m*)<br>/return ⊤ if there is a next operations sequence for *m* or one of its predecessors, ⊥ else/<br>2:    Visited ← Visited ∪ {*m*}<br>3:    **if** GetInstance(*m*) = ⊥ **then**<br>4:       **return** ⊤ /*m* has to be (re)initialized/<br>5:    **end if**<br>6:    **for all** *m'* ∉ Visited ∧ *m'* ∧ ∈ GetInCover(*m*) **do**<br>7:       **if** NextOpSeq(*m'*) **then**<br>8:          **return** ⊤<br>/there is a next instanciation of *m'*/<br>9:       **end if**<br>10:  **end for**<br>11:  **if** NextInCover(*m*) ≠ ⊥ **then**<br>12:    **return** ⊤<br>/there is a next input cover of the currently selected instance for *m*/<br>13:  **end if**<br>14:  **if** NextInstance(*m*) ≠ ⊥ **then**<br>15:    NextInCover(*m*)<br>16:    **return** ⊤ /there is a next instance for *m*/<br>17:  **end if**<br>18:  **return** ⊥ /there is no next instance of *m*/<br>19: **end procedure** |

## 3.4 Example

### 3.4.1 Find a meta-composition

Given a user request with knowledge K = {a}, goal G = {x, y, z}, and *N* the interaction network of meta-operations in Figure 2, Procedure 1 is used to find a meta-composition *C* within *N*.

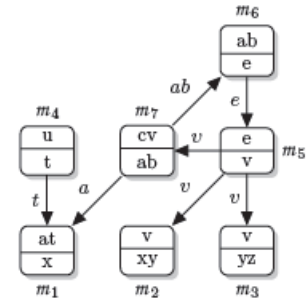

**Figure 2**. Example of an interaction network of meta-operations explored by step 1 to find a meta-composition.

First, $m_1$, $m_2$ and $m_3$ are the only meta-operations to supply at least a part of the goal. They provide initial nodes to explore *N*. The zero-length path {$m_1$} is considered at first. Since a is known and one of its input parameter, $m_1$ is selected for meta-composition. Path {$m_4$, $m_1$} is hold for further exploration. {$m_7$, $m_1$} is ignored because a is already known and $m_1$ does not need b. When considered, {$m_4$, $m_1$} is finally discarded because no input parameters of $m_4$ is known.

Then, zero-length path $\{m_2\}$ is considered before paths $\{m_5, m_2\}$, $\{m_6, m_5, m_2\}$ and $\{m_7, m_6, m_5, m_2\}$. Among those, only $\{m_6, m_5, m_2\}$ has a first meta-operation with a known input parameter, thus its nodes are selected as a meta-composition. Although $m_5$ is a predecessor of $m_7$, path $\{m_5, m_7, m_6, m_5, m_2\}$ is not hold for exploration, because it would be cycling at $m_5$.

At the end, the zero-length path $\{m_3\}$ is considered and discarded because no one of its input parameter is known. Then $\{m_5, m_3\}$ is considered. Its first meta-operation is already in the meta-composition as it is a suffix of an invocable path; thus $m_3$ is also added to the meta-composition. The resulting meta-composition is a sub-network of $N$ restricted to vertices $\{m_1, m_2, m_3, m_5, m_6, m_7\}$.

### 3.4.2 Refined the meta-composition

Given the meta-composition $C$, in Figure 3(left), resulting from the previous step, Procedure 5 is used to refine $C$. It guarantees that the refined meta-composition $C$ is only composed of instantiable meta-operations. The refined meta-composition is no more a sub-network of $N$. It is still an interaction network, but with slightly modified meta-operations. Compared to the meta-composition as found at the previous step, while $m_7$ is removed from meta-operations, $o_1''$ and $o_6'$ are removed from the underlying operations of their respective meta-operations $m_1$ and $m_6$. $m_1$ is replaced by $m_1'$. The resulting meta-composition is $\{m_1', m_2, m_3, m_5, m_6\}$.

Given the refined meta-composition resulting from the previous step, Procedure 2 is used to instantiate this refined meta-composition. As shown in Figure 3(right), there are two alternatives that cover the goal, one with $\{m_1', m_3\}$, the other with $\{m_2, m_3\}$. There are also two possible instances for $m_1'$, the operations $o_1$ and $o_1'$. Even if they are functionally equivalent, they might have different implementations, or belong to different Web services. Hence they are prone to have different non-functional properties. At the end, instantiated compositions are: $\{o_6, o_5, o_3, o_1\}$, $\{o_6, o_5, o_3, o_1'\}$ and $\{o_6, o_5, o_3, o_2\}$. Note that if no solutions are found in the FullSim network, the others similarity networks can be considered, depending on user's preferences.

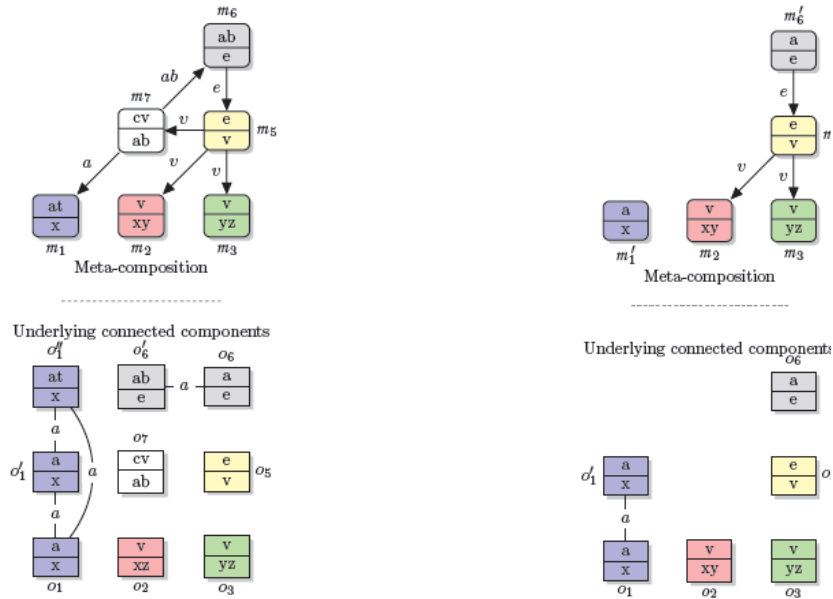

**Figure 3**. Left: Example of a meta-composition going to be refined by Procedure 5. Right: Example of a refined meta-composition used by Procedure 2 for instantiation.

**Procedure 5.** Refine meta-composition
1: *Grays* ← *C*
2: *WasMod* ← ⊤
3: **while** *Grays* ≠ ∅ ∧ *WasMod* **do**
4:     *WasMod* ← ⊥
5:     **for all** *m* ∈ *Grays* **do**
6:         *K'* ← ∅
7:         *K''* ← ∅
8:         **for all** *m'* ∈ Pred(*m*, *C*) **do**
9:             **if** *m'* ∈ *Grays* **then** /*gray* meta-operations are in *Grays*/
10:                *K''* ← *K''* ∪ Label(*m'*, *m*)
11:            **else if** *m'* *C* ∈ **then** \instanciable meta-operations are in *C* \ *Grays*\
12:                *K'* ← *K'* ∪ Label(*m'*, *m*)
13:            **end if**
14:        **end for**
15:        **for all** o ∈ *m* **do**
16:            **if** Input(o) ⊆ *K* ∪ *K'* **then**
17:                *Grays* ← *Grays* \ {*m*} \assert that *m* is instanciable\
18:                *WasMod* ← ⊥
19:            **else if** Input(o) ⊈ *K* ∪ *K'* ∪ *K''*
                   **then**
20:                *m* ← *m* \ {o} \assert that o is not invokable\
21:            **end if**
22:        **end for**
23:     **end for**
24: **end while**
25: *C* ← *C* \ *Grays* \remove gray meta-operations from meta-composition *C*\

## 4. Conclusion

In this paper we proposed a two-layered architecture to tackle the large scale and redundancy issue occurring in Web service composition synthesis. This architecture is based on network representations of the Web services space. A concrete layer is made of communities of similar Web services operations. It is realized by a similarity network of operations. An abstract layer is made of interacting meta-operations, each meta-operation being the representative of a community. It is realized by an interaction network of meta-operations. A graph-based search algorithm acts in two phases to explore the network architecture. In the first phase, it starts from the desired goal of a request for exploring the interaction network and retrieve a meta-composition. In the second phase, it starts from the meta-composition found during the first phase and explores the similarity network to replace the meta-operations by corresponding operations.

The main contribution of the proposed approach is to reduce drastically the search space as compared to previous graph based approaches. Indeed structuring the Web services operation space into communities allows dealing efficiently with the redundancy issue. Furthermore, similar operations can be easily substituted. This aspect is of great value because Web services are highly volatile.

However, our algorithm presents some limitations. It is unable to select the most valuable composition according to user criteria, without requiring an exhaustive instantiation of the meta-composition. Furthermore it does not consider subsumption relationships of ontological concepts in order to define operations similarity and it does not take advantage of the topological structure of the networks. Our future work will address these issues.